# Resource-constrained FPGA Design for Satellite Component Feature Extraction

Andrew Ekblad
Florida Institute of Technology
150 W. University Blvd
Melbourne, FL 32901
aekblad2019@my.fit.edu

Trupti Mahendrakar
Florida Institute of Technology
150 W. University Blvd
Melbourne, FL 32901
Tmahendrakar2020@my.fit.edu

Ryan White
Florida Institute of Technology
150 W. University Blvd
Melbourne, FL 32901
rwhite@fit.edu

Markus Wilde
Florida Institute of Technology
150 W. University Blvd
Melbourne, FL 32901
mwilde@fit.edu

Isaac Silver
Energy Management Aerospace
2000 General Aviation Drive
Hanger 101
Melbourne, FL
isaac@energymanagementaero.com

Brooke Wheeler
Florida Institute of Technology
150 W. University Blvd
Melbourne, FL 32901
bwheeler@fit.edu

*Abstract*—The effective use of computer vision and machine learning for on-orbit applications has been hampered by limited computing capabilities, and therefore limited performance. While embedded systems utilizing ARM processors have been shown to meet acceptable but low performance standards, the recent availability of larger space-grade field programmable gate arrays (FPGAs) show potential to exceed the performance of microcomputer systems. This work proposes use of neural network-based object detection algorithm that can be deployed on a comparably resource-constrained FPGA to automatically detect components of non-cooperative, satellites on orbit. Hardware-in-the-loop experiments were performed on the ORION Maneuver Kinematics Simulator at Florida Tech to compare the performance of the new model deployed on a small, resource-constrained FPGA to an equivalent algorithm on a microcomputer system. Results show the FPGA implementation increases the throughput and decreases latency while maintaining comparable accuracy. These findings suggest future missions should consider deploying computer vision algorithms on space-grade FPGAs.

## TABLE OF CONTENTS

## NOMENCLATURE

| | |
|---|---|
| DPU | = Deep learning Processor Unit |
| FPGA | = Field Programmable Gate Array |
| IP | = Intellectual Property |
| NCS2 | = Neural Computer Stick 2 |
| NMS | = Non-Max Suppression |
| PL | = Programmable Logic |
| PS | = Processing System |
| SOC | = System on Chip |
| SOM | = System on Module |
| YOLO | = You Only Look Once |

## 1. INTRODUCTION

Neural networks require extensive computational resources and are therefore generally slow and power consuming. Traditionally, spaceflight computers are too slow to run these algorithms. However, field programmable gate arrays (FPGAs) have been used on spacecrafts for a long time and are typically used for signal processing. Recent developments in space-grade FPGAs like Xilinx Versal have opened the door to using them for deep learning applications in space. This paper will cover a method for training and quantizing a neural network that can be deployed on a Xilinx FPGA. This is accomplished by using a Xilinx KV260 and Vitis AI to deploy the You Only Look Once (YOLO)v4 object detector for satellite feature extraction. The accuracies of the original trained model and quantized model is compared. The performance of the FPGA-deployed model is then analyzed, and its inference performance is compared to that of a similar model that runs on a Raspberry Pi 4B and an Intel Neural Compute Stick 2 (NCS2).

## 2. MOTIVATION AND PREVIOUS WORKS

Our previous work focused on-orbit servicing and active debris removal operations [1]. This project has two parts, an artificial potential field (APF) algorithm [2] for flightpath planning, and a machine vision algorithm paired with stereographic cameras for depth estimation. The machine vision algorithm is used to capture information about a non-cooperative spacecraft to allow the APF algorithm to plan a path to dock with it. Since the spacecraft is non-cooperative, a chaser satellite must approach a safe capture point, the body of the target satellite for example. It must also avoid fragile components like solar panels or antenna. Our current implementation uses an Intel NCS2 [3] and a Raspberry Pi on machine vision tasks [4].

Performance with this current implementation was significantly lowered that desired and is far from real time. Our current framerate is around 2 FPS, and it has proven a challenge to try improving this performance. This work is an attempt to try an alternative approach to increase inference speed compared to our current implementation.

## 3. YOLO

In recent years, computer vision has rapidly become more effective in many domains, from autonomous driving to automated feature extraction from satellite imagery to in-space guidance, navigation, and control (GNC) and beyond. At the core of these developments are deep convolutional neural networks (CNNs), most typically accelerated with graphics processing units (GPUs) [4]. These models have even exceeded human capabilities at certain vision tasks, such as classifying the ImageNet dataset [5]. Over the past 5-6 years, CNN optimized for edge computers have arisen, enabling at least some of these capabilities on small computers with low power consumption.

The work herein is part of a larger research program tasked with exploring how small, low-power (spaceflight-like) computers can employ computer vision to enable autonomous rendezvous with non-cooperative resident space objects, particularly satellites. This requires locating safe capture points on a satellite body as well as fragile components like solar panels or antennas. Safe flightpaths can then be planned to guide chaser satellites around these hazards and to capture points. The common computer vision task of object detection does just this: it aims to identify objects, localize them by drawing a tight bounding box surrounding them, and classify the type of object from a pre-determined set of classes. That means solar panels, satellite bodies, antennas, and thrusters.

While there are many object detectors, only single-stage detectors have a low enough computational footprint for deployment on spaceflight-like hardware with suitably high framerates. Single-stage detectors learn to estimate bounding boxes, classifications, and objectness scores (confidences) through a single forward pass of a neural network processing a single image frame. The most successful of these are the You Only Look Once (YOLO) [7] family of algorithms. Initially released in 2015, it has undergone several iterations, YOLO9000 (YOLOv2) and YOLOv3 from the same author and YOLOv4 [8] in 2020, each iteratively improving its accuracy by adjusting the neural architectures and tuning the hyperparameters of the model.

## 4. WHY USE AN FPGA?

Typically, CPUs, GPUs and application specific integrated circuits (ASICs) are used to perform machine learning inference. However, FPGAs can be used as an alternative. Both Intel and Xilinx have created tools that allowed users to run machine learning algorithms on their hardware. Intel has an FPGA AI suite [9] and Xilinx has Vitis AI [10] which uses Xilinx's Deep Learning Processing Unit (DPU). In their testing Xilinx's DPU can produce similar performance to an embedded GPU like the NVIDIA Jetson with lower power consumption [11].

*DPU IP*

Xilinx's Intellectual Property (IP) for running neural networks on their hardware is the DPU [12]. The DPU is a highly configurable IP that can be used for neural networks and is specifically designed with convolutional neural networks in mind. The highly configurable DPU can be used in larger complex designs and can be configured for the various area and performance requirements. The main parameter that defines size and performance of the DPU is the number of operations per clock it can handle. Operations per clock can range from 512 to 4096. The DPU IP can be configured to use up to 4 DPU cores in parallel in one design. The DPU IP is available on several of Xilinx's product lines, including Zynq UltraScale+, select Alveo devices and select Versal devices [10]. The DPU can also be used along with other components in an FPGA design allowing for the designer to add more functionality on the same chip.

*FPGA Flexibility*

As mentioned above a single FPGA can be used to integrate the DPU IP, but it can also be used to implement more functions. Additional functions could include inference pre and post processing, which could be used to accelerate the parts of the model that are not computed using the DPU (this is discussed more in the results section and future research area).

Another example would be an image sensor processing pipeline that would allow a reduction in latency from sensor to the memory that is shared with the DPU when compared to using other camera interfaces such as USB.

*FPGA vs Other Computing Devices*

Embedded GPUs are well-known for their use in machine learning because of their parallel architecture. Systems such as the NVIDIA Jetson [13] are good options for embedded machine inference. The market also has ASICS that are designed for machine learning inference such as Google's TPU [14] and Intel's NCS2 [3]. The main reason to use an FPGA instead of any of these devices really comes down to the flexibility of FPGAs—not just their ability to implement more than a single function but also that they can easily be updated after they are deployed.

## 5. VITIS AI WORKFLOW

Vitis AI is Xilinx's platform for quantizing, compiling, and running inference of a trained model on their FPGA platforms.

*YOLOv4 Model Training*

In this work, the model was trained using a modified version of TensorFlow YOLOv4 [15]. The modifications to the model are required because the DPU IP has a limited number of supported operations. Therefore, the neural network must be modified to contain only the supported layers. If the model contains unsupported layers, it will either offload unsupported instructions to the CPU or not run at all. Offloading operations to the CPU would cause a significant slowdown and the performance reduction would be unacceptable.

YOLOv4 requires two modifications that must be completed before training to prepare for use on the DPU. The first is to replace the mish activation function [16], defined in equation (1) and shown in Figure 1.

$$\text{mish}(x) = x * \tanh(\ln(1 + \exp(x))) \quad (1)$$

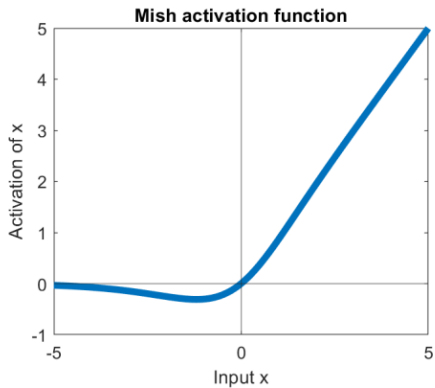

**Figure 1. Mish Activation Function**

with the leaky ReLU activation function [17], which is defined in (2)

$$\text{Leaky ReLU}(x) = \begin{cases} ax, & x < 0 \\ x, & x \geq 0 \end{cases} \quad (2)$$

and is shown in Figure 2 where "a" is a learnable parameter.

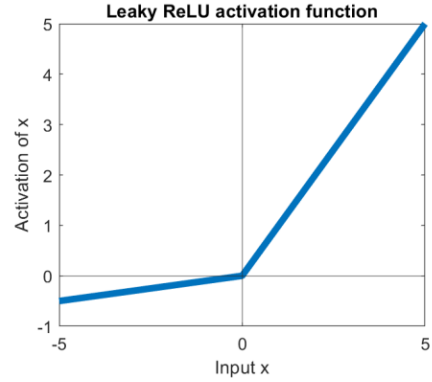

**Figure 2. Leaky ReLU with a = -0.1**

This change to the activation function is required because the DPU does not have support for the mish activation function. Computationally the leaky ReLU function is also considerably faster than the mish function. The change in activation functions results in similar behavior.

The second change is to reduce the maximum size of the max pooling kernel down to 8x8, because larger sizes are not supported by the hardware [12].

Once the modifications are made the model may be trained as normal using the any dataset that is compatible with YOLOv4. Specific details about training the model in this paper are left until the implementation section. Once the model is trained, it must be frozen then quantized so that it can run on the device.

*Model Quantization*

Model quantization allows the model to convert the weights and activations from single-precision floating-point (FP32) to 8-bit computations (INT8) with minimal loss in accuracy [10] Vitis AI provides libraries to quantize many common types of model frameworks. TensorFlow in this work. Model quantization in Vitis AI is done by using a set of unlabeled images from the dataset to analyze the distribution of activations [10]. Quantization has a small effect on the accuracy of the trained model [10]. Specific details of quantization in this work will be discussed in the implementation section below.

*Model Compiling*

After the model has been quantized, the model needs to be compiled for the specific DPU architecture. Once the model has been compiled it is possible to run the model directly on the DPU, the compiled model will automatically sperate what layers can run on the DPU and the CPU using the Vitis AI framework.

## 6. IMPLEMENTATION

*Training Dataset*

The training dataset that was used for this paper has been used extensively for other works where we have shown that it is capable of detecting satellites [18]. The dataset consists of 1,231 images that contain 7,971 annotations across 4 classes. The four classes contained in the dataset are body, solar, antenna, and thruster. Thrusters were not included in this work because we have found that they generally perform poorly with small image sizes. The dataset was split into two structures, validation, and training. The model was trained using the training set, and was quantized using the validation set.

The testing dataset is a set of images that was captured at the Florida Institute of Technology ORION facility [19], with the Kinematics Simulator. The Kinematics simulator uses a model satellite and a gantry system to emulate satellite movement in space. Information about the kinematics simulator to its full extent and can be found here [18]. The testing dataset was used to test the model on the FPGA.

*Model Selection*

YOLOv4 [8] was selected for this work because our previous works [4] used YOLOv5 [20] for object detection and YOLOv4 has a similar model architecture. Using a similar model to our previous work also allows this work to be more directly compared to what was achievable using YOLOv5 and Raspberry Pi + Intel NCS2 [4]. YOLOv5 was not chosen for this work, due to additional complexities in the implementation, however in future works we would like to explore deploying it on the FPGA.

*Device selection*

The device that was selected for this work was the Xilinx KV260 [21]. It is a device that is based on a Zynq UltraScale+ MPSoc. This family of Zynq UltraScale+ devices has a quad-core Arm Cortex-A53 processor and a dual-core Arm Cortex-R5F real time processor which make up the Processing System (PS). As well as 4GB of onboard RAM and Programmable Logic (PL) block that functions as the FPGA [22]. The combination of the PS, RAM and PL make up the System On Module (SOM). The KV260 uses this SOM on a carrier board with select hardware to make the evaluation board.

The KV260 kit was specifically selected because the intended use of the product is AI. This means it has support for Xilinx's DPU IP, and it also has native support for DPU on PYNQ [23], The significance of which is discussed in the next section.

*DPU on PYNQ*

PYNQ [24] enables FPGA acceleration without the need for VHDL or Verilog for development [24]. The typical design flow would be accomplished using synthesizable languages accompanied with complex device drivers. This process typically requires extensive knowledge of FPGA design and a considerable amount of time to implement correctly. PYNQ enables the reuse of PL circuits as overlays without the need to reconfigure/recreate hardware and drivers for each project [24]. PYNQ also allows the user to create their own designs if the existing overlays do not meet their required needs, which requires some FPGA knowledge. In this design, the provided overlays were able to meet the needs of this research. The main element of the paper was Xilinx's DPU IP which makes use of Vitis AI. PYNQ includes an overlay for the DPU IP called DPU on PYNQ [23] This implementation has predeveloped PL hardware and drivers for the DPUCZDX8G configured as B4096 [23]. This means that all we needed to make use of the DPU is to import it into the design using Python function calls.

*Running the Model*

After the model has been trained, quantized, and compiled, it is now possible to run it directly on the hardware. Neural network models are typically broken into three parts, preprocessing, the neural network itself, and the post processing.

Preprocessing is the stage where the input image is converted into a form that can be accepted as the input for the neural network. This model has an input size of 416x416x3, which means 416 width, 416 height and 3 channels for color (RGB). An image that is not 416x416 will need to be resized to 416x416. In this work the image is resized by scaling the image down and bars are added to preserve the original aspect ratio. The preprocessed image is then given to the input of the DPU, and then DPU will run inference on the image. Post processing takes the output from the DPU, which must be processed before boxes can be drawn. The postprocessing method used with YOLO requires throwing away low confidence predictions as well as non-max suppression to eliminate duplicate bounding boxes [21]. After postprocessing, bounding boxes are drawn back onto the original image and the inference for a frame is complete.

The DPU is capable of running the neural network itself, but both for this work pre and post processing were implemented in software and ran on the CPU. CPU post processing is very slow, which is discussed in the next section.

## 7. RESULTS

Data used for testing was a small section of pictures extracted from videos that were taken using the Kinematics Simulator at the Florida Institute of Technology ORION facility [19]. These videos were captured using a GoPro, frames at a rate of 1 FPS were extracted and labeled. The images are of the test satellite in the lab which is shown in Figure 3.

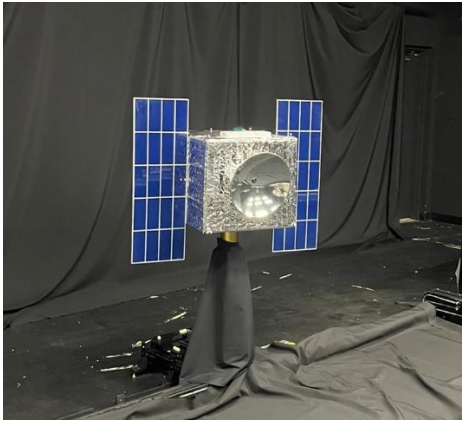

**Figure 3. Picture of Test Satellite**

The satellite has two solar panels an antenna and a thruster attached to the body. The video that was captured was of a simple satellite rotation around a single axis. From the captured video 40 frames were extracted and labeled to be used as the testing set. Since the satellite in the lab is used for testing, images of it are not included in any of the model training or validation sets. It was only once that model was trained that the model's accuracy would be tested on the satellite in the lab. This separation ensures that when the model looks at the satellite in the lab it has no prior knowledge about it. It should then be a good interpretation of how the model can generalize to satellites that it has not seen before. Since this satellite in the lab is physical and most of the images in the testing and validation set are computer generated it is expected that the performance on these images should be worse than the testing and validation set.

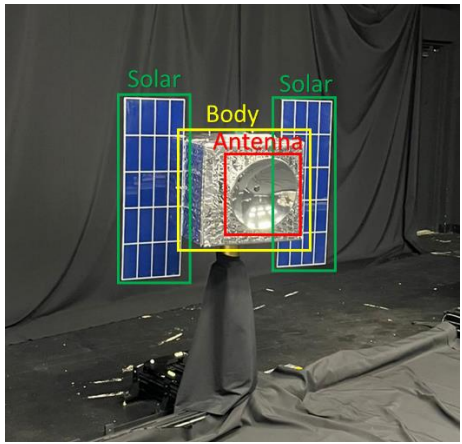

**Figure 4. Example of Bounding Boxes for Satellite**

Figure 4 extracts the components from the picture in Figure 3 and labels the class of each component.

*Training Results*

The model was trained using the modified version of YOLOv4 described in Vitis AI workflow section.

Figure 5 shows the loss from training, vs the number of epochs.

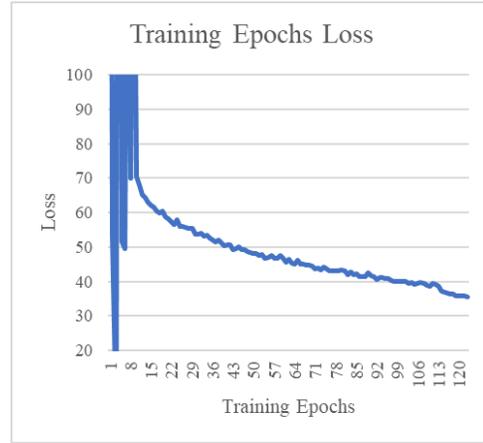

**Figure 5. Training Epochs Loss**

The trend in the training loss shows that the model accuracy could have been improved if the model was trained for longer. If we spent more time on model training we expect that the model's accuracy should close to our previous works [26].

After the model was trained, it needed to be tested. These tests were done on the testing set. The results for each class's average precision are shown in Table 1. The average precision for this model is very low, but previous works with the same training dataset and similar testing sets have shown better results [26]. Previous work compares YOLOv5 and faster R-CNN and was able to achieve an mAP@0.5 of 53.05% [26] using YOLOv5 while this model was only able to achieve an mAP@0.5 of 9.2%.

**Table 1. Original Trained Model Average Precision**

| Class | Average Precision |
|---|---|
| Solar | 02.31% |
| Body | 21.29% |
| Antenna | 13.31% |

This poor performance is not due to the use of the modified YOLOv4 model. Since this paper is focused on the FPGA implementation the performance of the trained model was not as much of a concern as the difference between the trained model and the quantized model.

*Quantization results*

The trained model was quantized using the quantization functions provided by Vitis AI. Once the model was quantized, it was tested on the same testing images with the same parameters as the original model. The results from this testing are shown in Table 2. The results when compared to Table 1 show similar average precision for both solar and antenna, however average precision significantly reduced for the body class.

**Table 2. Quantized Model Class Average Precision**

| Class | Average Precision |
|---|---|
| Solar | 02.21% |
| Body | 10.98% |
| Antenna | 12.15% |

The mean Average Precision (mAP) was measured at an IOU threshold of 0.5 only, these results are shown in Table 3. Similar to the average precision, the quantized model showed reduced mAP. If the model were trained longer, the difference between the original and the quantized model would likely be reduced.

**Table 3. Original and Quantized Model mAP @ IOU 0.5**

| Model | mAP@.5 |
|---|---|
| Quantize | 6.3% |
| Original | 9.2% |

*FPGA Inference Results*

After quantization the model was compiled for the hardware, the quantized model and the compiled model should be functionally the same. Therefore, no quantitative values about model accuracy were tested on the FPGA. Images of inference results are included in this paper to show functionality.

Figure 6Figure 1 shows the detections on the test dataset from a front view. Figure 7 shows detection on the test dataset from a rotated view. Figure 8 shows detections when the satellite is rotated 180°.

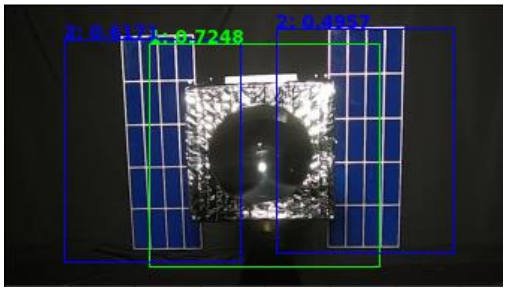

**Figure 6. FPGA detection results on satellite front**

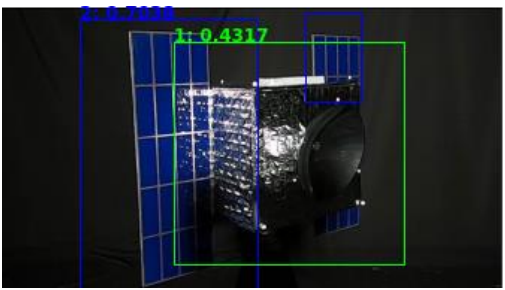

**Figure 7. FPGA detection results on satellite rotated**

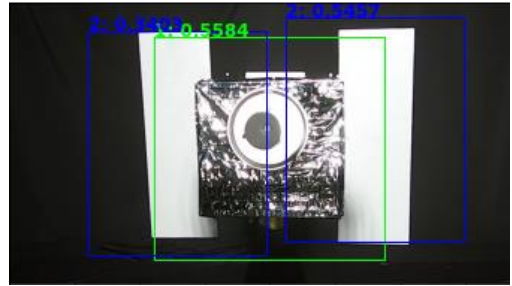

**Figure 8. FPGA detection results on satellite back**

The results from the FPGA show that the model is still functional when it runs on hardware. The detections from the bounding boxes are not well fitted to the object, but this can be attributed to the poorly trained model's generalization to the lab satellite, and not the fault of the FPGA. Figure 9 shows the detection from the same model, but on an image from the validation set. The bounding box confidence and locality are significantly improved showing that the model is able to find localize and categorize objects.

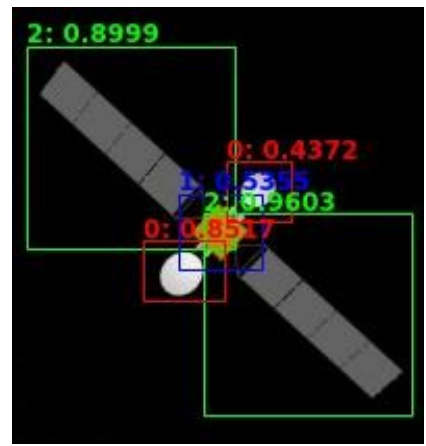

**Figure 9. FPGA detection using pretrained COCO dataset**

Since model inference speed is an important part of this work FPGA inference speed was tested next.

*FPGA Inference Framerate and Latency*

Inference speed was tested by first preprocessing all of the images from the testing dataset and loading them into an array. The preprocessing step resizes the image using letterboxing to scale the image size to 416x416 without changing the aspect ratio. Each preprocessed value in the array was passed to the DPU, the output from the DPU passed through post processing and output for a single image was obtained. The implementation in this work has both pre and post processing implemented on the CPU. The time it took to run this process was measured for 3 samples and the results are tabulated in Table 4.

**Table 4. FPGA Inference Latency**

| Sample | Latency ms |
|--------|------------|
| 1      | 272.1 ms   |
| 2      | 271.6 ms   |
| 3      | 266.9 ms   |

These results show that the latency from image input to image output is about 270ms.

Next, the frame rate was tested. Framerate is a way to measure inference throughput. Testing framerate was broken down into three tests to measure performance of the neural network, post processing and both.

These tests were run by preprocessing and saving all 40 testing images into memory. Each element was iterated through and given to the DPU for inference, the output from the DPU was then saved into another array. Frames Per Second (FPS) was calculated from the average of the total time it took the DPU to process the 40 images. Post processing was done in a similar manner, except the input array was the output DPU test. Finally average frame rate was measured from the input of the DPU to the output of post processing. The results from these tests are tabulated in Table 5.

The DPU was able to achieve a throughput of 9.3 FPS, but the post processing throughput was only 6.5 FPS. The average throughput is even lower than the average of the two throughputs.

**Table 5. Hardware Performance for Different Elements**

| Tested part                | Frame Rate |
|----------------------------|------------|
| DPU Throughput             | 9.3 FPS    |
| Post Processing Throughput | 6.5 FPS    |
| Average Throughput         | 3.8 FPS    |

The difference in average throughput can be explained because the neural network calculations are handled by the DPU which makes it very fast. However, model post processing which takes most of the time. Model post processing is the stage where all of the output predictions need to be checked to see what their confidence is, the output predictions of the model has 3549 parameters and each of these has a confidence value that needs to be checked. This is a slow process because there are many results that need to be evaluated. Many of the filtered results will get discarded since only results that have a confidence value greater than the threshold are kept. This computation causes the CPU to spend time checking many output values that will ultimately be discarded.

*Areas of improvement*

Since the model output filtering is a slow process, there are several possible solutions to speed up filtering.

First since each output from the model has independent data, meaning data from one iteration does not depend on data from another iteration. Therefore, the post processing operation can be parallelized this parallelization can be done on the CPU by filtering using multiple processing cores or it could be implemented in the PL on the FPGA using custom filtering logic. Each option has their own advantages and disadvantages. Implementing on the FPGA could possibly offer greater parallelization but would require more complicated logic. Implementing on the CPU would not require implementation in hardware, but could require more complicated software. It may also not be as fast as implementing in the PL.

The second strategy that could improve performance is using pipelining. The current implementation uses one hardware unit at a time, meaning the DPU performs an operation then the CPU performs an operation. During this time the DPU is sitting idle. It is possible to pipeline the operations such that the DPU and the CPU are both working on something. In theory this should increase the maximum throughput to the slowest operation which is currently the post processing.

*Performance Comparison with Microcomputer Devices*

Our previous works have involved performing similar tasks using an Intel NCS2 and a Raspberry Pi 4 [4]. We achieved a frame rate of around 2 fps with varying latency with that hardware. With the work presented in this paper, we achieved consistent latency and 3.8 FPS. While this might not seem like an improvement, because this device is an FPGA and the post-processing was found to be the bottleneck, it should be possible to speed up inference by implementing post processing on the FPGA alongside the DPU.

## 8. SUMMARY

Using Vitis AI, and a modified YOLOv4 model an FPGA was able to run inference and extract features of satellites. The results show that this work is capable of running inference with YOLOv4 at 3.8 frames per second. The conversion of the model to run on the FPGA showed a reduction in accuracy when compared to the original model training. This work was able to meet the same performance as an Intel NCS2 and Raspberry Pi from our previous works. Since the new implementation was implemented on an FPGA, it is believed that implementing the slow CPU instructions on the FPGA and adding pipelining could significantly improve performance. Future works will involve exploring possible methods to accelerate post-processing and increase throughput

## ACKNOWLEDGEMENTS

The work on this study was supported by the AFWERX STTR Phase II contract FA864921P1506. Additional conference funding support was provided by N000142012669 Inspiring Students to Pursue U.S. Navy

STEM Careers through Experiential Learning grant. The authors thank Mackenzie Meni for proofreading the paper.

## BIOGRAPHY

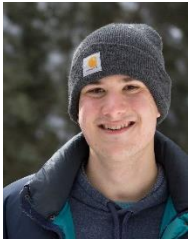
*Andrew Ekblad* completed his B.S. in Electrical Engineering from Florida Institute of Technology and is currently pursuing his M.S. in Electrical Engineering from Florida Institute of Technology. He is interested in high performance embedded computing systems.

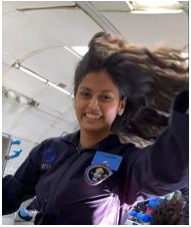
*Trupti Mahendrakar* received a B.S. in Aerospace Engineering from Embry-Riddle Aeronautical University, Prescott, Arizona in 2019 and an M.S, in Aerospace Engineering from Florida Institute of Technology, Melbourne. She is currently a Ph.D. student at the Florida Institute of Technology, Melbourne. Her current research includes the implementation of machine vision algorithms to enhance on-orbit service satellite operations, optimization of cold gas thruster design, and implementation of robotic manipulators for satellite refueling.

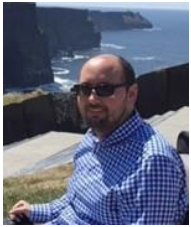
*Dr. Ryan T. White* is an Assistant Professor of Mathematics at Florida Tech and Director of the NEural TransmissionS (NETS) Lab. His research focuses on computer vision, physics-inspired machine learning, synthetic data generation, and probability. His research focuses on computer vision, physics-inspired machine learning, synthetic data generation, and probability. His work includes multidisciplinary projects on in-space use of computer vision to support on-orbit proximity operations, medical data analytics, glaciology, and geoinformation systems. Dr. White earned a Ph.D. from Florida Tech and joined the Florida Tech faculty in 2015.

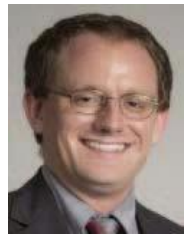
*Dr. Markus Wilde* is an Associate Professor for Aerospace Engineering at Florida Tech and Director of the ORION Lab. His research focus lies on experimental studies of spacecraft and aircraft control systems. Dr. Wilde received his M.S. and Ph.D. in Aerospace Engineering at TU Munich, Germany. He was accepted into the NRC Research Associateship Program in 2013, as postdoctoral associate at the Spacecraft Robotics Laboratory at the Naval Postgraduate School. In 2014, he joined the Florida Tech faculty.

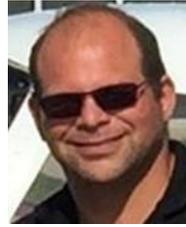
*Dr. Isaac Silver* is the CEO of Energy Management Aerospace. He earned his Ph.D. from Florida Tech in Space Sciences. He also holds a B.S. in Astronomy and Astrophysics from Florida Tech. He's an Airline Transport Pilot (Airplane Multi-Engine Land), Commercial Pilot (Airplane Single-Engine Land and Sea), Gold Seal Flight Instructor (Single and Multi-Engine), Instrument Pilot (Airplane) and Ground Instructor (Advanced). He has 17,500 hours of flight time with more than 4,000 hours as an instructor. Aircraft include DA-10, DA-20, DA-50/900, L-1329, BE-350, BE-400, IAI-1124, Learjet (24/25/31/35), and L39.

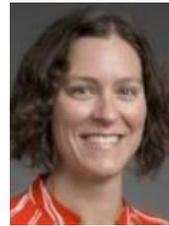
*Dr. Brooke Wheeler* is the Director of the Teaching Assistant Seminar and Assistant Professor of Aviation Sciences at Florida Tech. She earned her Ph.D. in Ecology from the University of North Carolina at Chapel Hill. She was a Postdoctoral Fellow in the Thompson Writing Program at Duke University, teaching and learning about using writing in STEM courses, and helping to run the Writing Studio. At Florida Tech, Dr. Wheeler teaches quantitative research methods, aviation research, and aviation sustainability. Her research includes electric aircraft, aviation sustainability, flight training, safety management systems, and how to grow plants on Mars.